\documentclass{article}

\usepackage[nonatbib,preprint]{nips_2018}

\usepackage[utf8]{inputenc} % allow utf-8 input
\usepackage[T1]{fontenc}    % use 8-bit T1 fonts
\usepackage{hyperref}       % hyperlinks
\usepackage{url}            % simple URL typesetting
\usepackage{booktabs}       % professional-quality tables
\usepackage{amsfonts}       % blackboard math symbols
\usepackage{nicefrac}       % compact symbols for 1/2, etc.
\usepackage{microtype}      % microtypography
\usepackage{amsmath,amssymb,cite,graphicx,color}

\title{Fairness GAN}

\author{
  Prasanna Sattigeri, Samuel C.\ Hoffman, Vijil Chenthamarakshan, and Kush R.\ Varshney\\
  IBM Research\\
  Yorktown Heights, NY 10598 \\
  \texttt{\{psattig@us.,shoffman@,ecvijil@us.,krvarshn@us.\}ibm.com} \\
}

\begin{document}

\maketitle

\begin{abstract}
In this paper, we introduce the Fairness GAN, an approach for generating a dataset that is plausibly similar to a given multimedia dataset, but is more fair with respect to protected attributes in allocative decision making.  We propose a novel auxiliary classifier GAN that strives for demographic parity or equality of opportunity and show empirical results on several datasets, including the CelebFaces Attributes (CelebA) dataset, the Quick, Draw!\ dataset, and a dataset of soccer player images and the offenses they were called for. The proposed formulation is well-suited to absorbing unlabeled data; we leverage this to augment the soccer dataset with the much larger CelebA dataset.  The methodology tends to improve demographic parity and equality of opportunity while generating plausible images.
\end{abstract}

\section{Introduction}
\label{sec:intro}

Automated essay scoring in high-stakes educational assessment \cite{Shermis2014,Perelman2014}, automated employment screening based on voice and video \cite{Shahani2015,Chandler2017}, and automated sports refereeing from wearable sensor data \cite{Ghafourifar2017} are all extant examples of decision making supported by machine learning algorithms with multimedia inputs.  They all raise concern about perpetuating and scaling unwanted discrimination that is present in historical data for implicit or explicit reasons. 

Unwanted discrimination against groups defined by protected attributes such as race, gender, caste and religion that directly prevents favorable (economic) outcomes such as being hired, paroled or given a loan has always been a problem in human decision making \cite{VarshneyV2017}, but has come to the fore as such decision making is being shifted from people to machines \cite{WilliamsBS2018}.  The study of such \emph{allocative harm} has spurred a recent flurry of research in the data mining and machine learning literature.  It is now well-known that there are three main classes of intervention to introduce fairness into supervised machine learning pipelines \cite{hajian2013simultaneous,ShaikhVMVRW2017}: pre-processing \cite{kamiran2012data,hajian2013methodology,zemel2013fairrepresentations,ruggieri2014using,CalmonWVRV2017}, in-processing \cite{kamishima2011fairness,fish2016confidence, zafar2017fairness,AgarwalBDL2017}, and post-processing \cite{hardt2016equality}.  

Much of the allocative fairness literature focuses on structured data. In this work, we examine pre-processing and find that existing approaches have not been applied to multimedia data and are not scalable to the high dimensionality that comes with such data.  Moreover, the subset of existing pre-processing methods that generate a new dataset by editing feature values are  unlikely to produce realistic pre-processed datasets when applied to multimedia data even if they could scale.  To address these limitations, we propose a new debiasing approach based on generative adversarial networks (GANs) \cite{Goodfellow2017}.  Specifically, we use the auxiliary classifier GAN (AC-GAN) as the starting point for our proposal \cite{OdenaOS2017}.  This is the first application of GANs to algorithmic fairness in our knowledge.  We name this methodology the Fairness GAN.

Recent contributions examining fairness from the perspective of adversarial learning include references \cite{EdwardsS2016,BeutelCZC2017,ZhangLM2018,MadrasCPZ2018}.  These methods are similar to our proposed approach by including a classifier trained to perform as poorly as possible on predicting the outcome from the protected attribute, but are different from our proposed method in two key ways.  First, they are not intended to create a releasable new dataset that plausibly approximates a given original biased dataset but with the the traces of discrimination cleansed as much as possible; they are intended for learning fair latent representations that are not in the same space as the given original, which restricts the transparency of the transformation and the flexibility in the use of the dataset.  As such, they are not GANs with a real/fake discriminator component that aims to generate realistic samples.  Second, they are applied to low-dimensional structured data (Adult and Diabetes UCI datasets, Heritage Health Kaggle dataset) or word embeddings and are not deep.  

The method of \cite{EdwardsS2016} provides fairness in the sense of demographic parity; \cite{BeutelCZC2017} allows for fairness in the sense of equality of opportunity; and \cite{ZhangLM2018} further allows for fairness in the sense of equality of odds.  Herein, we propose versions of the Fairness GAN for the first two of these definitions and leave equality of odds for future work. 

The existing methods \cite{EdwardsS2016,BeutelCZC2017,ZhangLM2018,MadrasCPZ2018} can handle the case when the protected attribute is known only for a subset of samples and \cite{MadrasCPZ2018} additionally considers transfer learning.  In our work, we consider a similar case that is relevant because we are attempting to generate plausible signals: the dataset labeled with outcomes and protected attributes is small, but there exists a much larger multimedia dataset with similar features but no outcomes.  Here our solution is to augment the dataset for training the generator with samples from the larger dataset.

Reference \cite{CalmonWVRV2017}, among others, points to the need for controlling both group discrimination and individual sample distortion.  We explicitly pursue group discrimination control using the auxiliary classifier that is made to be as bad as possible.  The formalism of GANs naturally controls distortion to individual samples through the main discriminator trying to tell real and fake samples apart, and we do not need any further explicit machinery for individual fairness.

There is ongoing debate whether treatment parity is allowed in algorithmic decision making. As discussed in \cite[Section 3.III]{CalmonWVRV2017}, it is possible to either use the protected attributes in the final classifier or to suppress them.  Using them in a group-conditional decision is a form of treatment parity.  Reference \cite{ZafarVGGW2017} states that ``the use of group-conditional decision making systems is often prohibited'' whereas \cite{LiptonCM2018} ``call[s] this premise into question.''  In this work, we use the protected attributes only during training and not during testing; however, since our approach is pre-processing, it is not difficult to allow for it.

The GAN literature includes approaches for conditional manipulation of multimedia data \cite{PerarnauWRA2016}, unpaired translation \cite{ZhuPIE2017}, and their combination \cite{LuTT2017}, which have some relationship to the method proposed herein.  One differentiating aspect of our work is that we do not have a collection of samples from a target distribution that we are trying match.  Also, we are never in position to manually manipulate or change the value of some descriptive attribute of the multimedia data, and in fact our work right now is limited to the setting in which all predictive features are multimedia ones: there are no other metadata-like features. Additionally, in the fairness setting, there are objective measures of quality, demographic parity or equality of opportunity, that are defined over a collection of samples and cannot be judged on a single sample alone.

\textbf{Datasets.} We experiment with several datasets, including the CelebFaces Attributes (CelebA) dataset \cite{LiuLWT2015}, a dataset of images of soccer players \cite{Silberzahn2017}, and the Quick, Draw!\ dataset of hand-drawn sketches.\footnote{https://github.com/googlecreativelab/quickdraw-dataset} In CelebA, we treat the attractiveness attribute as the allocative outcome variable.  This attribute certainly has shortcomings, but is the only one in this dataset and more generally among people image datasets commonly used in the GAN literature that relates to a human decision of some sort of quality and is not simply descriptive, such as being bald, wearing glasses, or having a pointy nose.  We consider two different sets of protected attributes: gender, which is given in the dataset, and a binary skin tone label that we annotate using the Fitzpatrick skin type scale.  In the soccer dataset, the allocative decision is yellow cards and red cards.  The protected attribute is a skin tone category averaged from two annotators given with the dataset. In Quick, Draw!, we focus on the `power outlet' category of sketches.  The allocative decision is whether the sketch was judged to be a power outlet (a task equivalent to automatic test scoring).  The protected attribute is the country of origin of the submitter (Canada or Great Britain).

\textbf{Implementation.} In order to make the Fairness GAN effective, we adapt several recent techniques including conditional batch normalization \cite{dumoulin2016learned}, the projector discriminator architecture \cite{MiyatoK2018}, spectral normalization regularization \cite{MiyatoKKY2018}, and hinge adversarial loss \cite{2017arXiv170502894L}. 

\section{AC-GAN Background}
\label{sec:acgan}

The AC-GAN was recently developed to improve the training of GANs for image synthesis when the images come with a class label $C$ such as `monarch butterfly,' `daisy,' and `grey whale' \cite{OdenaOS2017}.  As in all GANs, the AC-GAN has a discriminator $D(\cdot)$ and a generator $G(\cdot)$ that work against each other, and are trained using so-called `real' samples distributed according to $X_\text{real}$.  Every generated sample is a function of a noise realization $z$ and additionally a class label realization $c$, i.e.\ $X_\text{fake} = G(c,z)$.  The objective functions for training the generator and discriminator are composed of the typical GAN objective, the log-likelihood of the correct source $S \in \{\text{real},\text{fake}\}$: 
\begin{equation}
	L_S = E[\log P(S = \text{real} \mid X_\text{real})] + E[\log P(S = \text{fake} \mid X_\text{fake})]
\end{equation}
as well as the log-likelihood of the correct class:
\begin{equation}
	L_C = E[\log P(C = c \mid X_\text{real})] + E[\log P(C = c \mid X_\text{fake})].
\end{equation}
The discriminator maximizes $L_S + L_C$ and the generator minimizes $L_S - L_C$.

\section{Fairness GAN Formulation}
\label{sec:fairnessgan}

The proposed Fairness GAN builds upon the AC-GAN. We use the same notation for the most part with $X$ as the image or other multimedia signal; the differences are as follows.  First, instead of using $C$ to indicate a class label, we use $C$ to denote the protected attribute label such as gender or caste. Second, we have an additional outcome variable $Y$ which is the allocative decision such as hiring or loan approval.

The objective of the Fairness GAN is to take a given real dataset $(C_\text{real}, X_\text{real}, Y_\text{real})$ and learn to generate debiased data $(X_\text{fake}, Y_\text{fake})$ such that the joint distribution of the features and outcome of the generated data (conditioned on the protected attribute) is close to that of the real data while yielding allocative decisions that have either demographic parity or equality of opportunity. Ideally, the outcome produced by the data generator would be independent of the conditioning protected attribute; we pursue this ideal by reversing the motive of the AC-GAN and introducing an auxiliary classifier trained to predict outcome from protected attribute as poorly as possible.  

The generator of the Fairness GAN produces both the features and outcome variables: $(X_\text{fake}, Y_\text{fake}) = G(c,z)$.  It contains two variations of the log-likelihoods of the correct source, one pair for joint $(X,Y)$ samples with source variable $S_J$:
\begin{align}
%\min_{g,h} \max_d\, 
L_{S_{J}}^R &= E[\log P(S_{J}=\text{real}\mid X_\text{real}, Y_\text{real})]\\
L_{S_{J}}^F &= E[\log P(S_{J}=\text{fake}\mid X_\text{fake}, Y_\text{fake})],
\label{eq:source_loss_j}
\end{align}
and one pair for multimedia $X$ features alone with source variable $S_X$:
\begin{align}
%\min_{g,h} \max_d\, 
L_{S_{X}}^R &= E[\log P(S_{X}=\text{real}\mid X_\text{real})]\\ 
L_{S_{X}}^F &= E[\log P(S_{X}=\text{fake}\mid X_\text{fake})].
\label{eq:source_loss_x}
\end{align}

We include a pair of class-conditioned losses to add structure to the GAN and help with training and generating plausible images.  These objective functions are the same as in the AC-GAN:
\begin{align}
%\min_{g,h} \max_d\, 
L_{C}^R &= E[\log P(C=c \mid X_\text{real})]\\ 
L_{C}^F &= E[\log P(C=c \mid X_\text{fake})].
\label{eq:class_loss}
\end{align}

Finally, for fairness, we include a pair of losses to encourage demographic parity:
\begin{align}
%\min_{g,h} \max_d\, 
L_{DP}^R &= E[\log P(C=c \mid Y_\text{real})]\\
L_{DP}^F &= E[\log P(C=c \mid Y_\text{fake})].
\label{eq:fair_loss}
\end{align}

The discriminator maximizes: 
\begin{displaymath}
L^D = L_{S_{J}}^R + L_{S_{J}}^F + L_{S_{X}}^R + L_{S_{X}}^F+ L_C^R + L_{DP}^R,
\end{displaymath}
and the generator minimizes:
\begin{displaymath}
L^{G}_{DP} = L_{S_{J}}^F + L_{S_{X}}^F - L_C^F + L_{DP}^F.
\end{displaymath}

For equality of opportunity, we take a cue from \cite{BeutelCZC2017,ZhangLM2018} and activate the fairness loss on the generator only for samples where $Y_{fake} = 1$. In the binary case, where $Y_\text{fake} \in [0,1]$, the generator minimizes:
\begin{displaymath}
L^{G}_{EO} = L_{S_{J}}^F + L_{S_{X}}^F - L_C^F + Y_\text{fake}L_{DP}^F.
\end{displaymath}

\section{Data Description}
\label{sec:data}

\subsection{CelebA}
CelebA is commonly used in deep learning research and consists of 202,599 color images of the faces of celebrities downloaded from the internet, cropped and resized to 64 pixels by 64 pixels \cite{LiuLWT2015}.  The images come with 40 binary attributes annotated by a ``professional labeling company'' that is further described in \cite{BohlenCS2017} as ``a group of 50 paid male and female participants, aged 20 to 30, and recruited from mainland China during a 3 month development phase.'' 

One of the 40 attributes is \emph{male} and another is \emph{attractive}.  In one set of experiments, we use \emph{male} as the protected attribute $C \in \{0,1\}$ without further delving into the social construction of gender or commenting on why the attribute is named as it is.  In another set of experiments, we use skin tone as the protected attribute.  CelebA contains multiple images of the same celebrity; we manually annotated one image each of the 10,177 unique celebrities and propagated the annotation to the rest of the images.  We used the Fitzpatrick skin type scale to do the annotation, with types I (ivory), II (beige), and III (light brown) categorized as $C = 0$ and types IV (medium brown), V (dark brown), and VI (very dark brown) categorized as $C = 1$.

We use \emph{attractive} as the outcome decision $Y \in \{0,1\}$ and assume it represents the labeler's judgment on the celebrity's attractiveness.  Several concerns with this attribute are presented by \cite{BohlenCS2017}, but we treat it as an allocative decision just like a hiring decision or a decision to accept an individual into a program.  

\subsection{Soccer (Many Analysts, One Dataset)}
We also consider a second dataset of images of people: soccer players from European professional leagues along with a record of their cautionable and sending off offenses.  This dataset was originally assembled for a unique crowdsourcing experiment testing whether different statisticians will find evidence of racial discrimination in the calling of offenses by referees \cite{Silberzahn2017}. 

The players have been labeled according to skin tone by two annotators, Lisa and Shareef, with five possible values \{0 (very light skin), 0.25 (light skin), 0.5 (neither light nor dark skin), 0.75 (dark skin), 1 (very dark skin)\}. We count players with average annotation value less than 0.5 as light and players with average greater than or equal to 0.5 as dark, and use this as the protected attribute $C$. 

The dataset contains counts of yellow cards, second yellow cards, and straight red cards given to the player.  We aggregate the count of all of these offenses over all matches and set $Y = 0$ for players with more than 0.12 offenses per match and set $Y = 1$ for players with less than or equal to 0.12 offenses per match.

The images are a mix of action shots and posed profile pictures from which we extract faces using the pre-trained Viola-Jones face detector and scale them to 64 pixels by 64 pixels so that they are similar to CelebA images.  The face detector does not find any faces in 5.3\% of the images, without any discernible bias related to $C$ or $Y$.  We drop these samples, yielding 1501 total samples in the dataset.  In images that the detector finds more than one face, we manually select the main player. We note that the profile image of the player is not the direct basis of individual yellow cards and red cards given by the referee.

\subsection{Quick, Draw!}
The final dataset we consider is not images of people, but sketches drawn by people. The Google Quick, Draw!\ dataset contains 50 million quickly drawn sketches of objects from 345 categories, such as `asparagus,' `swingset,' `The Mona Lisa,' and `zebra.' 

The sketches are captured as vectors of pen movements and also released as 28 pixel by 28 pixel bitmap images.  Unfortunately, it was not clear whether the correspondence between the image and the metadata was maintained in the released bitmap images. So, we created our own single-channel bitmap images of 64 pixel by 64 pixel size from the vector representation. Each sample is labeled by the country of the user that submitted the sketch. We use this variable as the protected attribute $C$. As the allocative decision $Y$, we use the binary assessment of the quality of the sketch (essentially a performance evaluation that could potentially be used for evaluating candidates for employment or admission to educational institutions).  This quality assessment, the variable named `recognized,' seems to be the output of an automated decision making model, but that is irrelevant to our work and could just as easily have been a human judgment.

We focus on the category `power outlet' as it is known to have differential recognition performance on submitters from different countries: poor for Great Britain and good for Canada and the United States \cite{TailSpectrum2016}.  We use a binary $C$ with Great Britain and Canada since the two countries have a similar number of samples for power outlet sketches.

\section{Implementation Details}
\label{sec:implementation}

The architecture of the generator and discriminator are based on the network structures in \cite{MiyatoKKY2018,MiyatoK2018,gulrajani2017improved}. The image generation path of the generator transforms the noise vector into image $X$ using a linear layer followed by 4 upsampling ResNet blocks \cite{HeZRS2016}. To generate class-conditional images, conditional batch normalization is employed \cite{dumoulin2016learned}. The same noise vector is passed through 2 dense layers to generate the outcome variable $Y$. ReLU functions are used as the activation functions in the intermediate layers for both the image and outcome generation paths. Meanwhile, $\tanh$ activation is used as the last layer for both paths. 

The discriminator provides 4 outputs: probability distribution over sources of the joint samples, probability distribution over sources of the image samples, probability distribution over classes conditioned on image samples and probability distribution over classes conditioned on outcome samples: $P(S_{J}\mid X, Y)$, $P(S_{X}\mid X)$, $P(C\mid X)$, $P(C\mid Y) = D(X,Y)$. The first three outputs share a common network $\phi(.)$ that transforms $X$. $P(S_{X}\mid X)$ and $P(S_{X}\mid X)$ are obtained using independent linear layers over $\phi(X)$. $P(C\mid Y)$ is obtained by passing $Y$ through $2$ dense layers with ReLU activation function. 

We use an architecture similar to projector discriminator \cite{MiyatoK2018} to obtain $P(S_{J}\mid X, Y)$. First, the outcome variable $Y$ is embedded in a space with same dimensionality as $\phi(X)$. This is followed by combining the embeddings of $X$ and $Y$ by inner product based interaction ($f(X,Y) = Y\mathbf{V}_{Y}\phi(X) + \mathbf{V}_{X}\phi(X)$) instead of the common approach of concatenation. 

In the case of the small cardinality soccer dataset, we leverage the much larger CelebA dataset to aid the learning of the generator. Skin tone is the protected attribute $C$ for both datasets, but CelebA is unlabeled with respect to the yellow card and red card outcome $Y$. To overcome the lack of outcome labels in CelebA, we tweak the discriminator loss to:
\begin{displaymath}
L^D = L_{S_{J}}^{R^{lab}} + L_{S_{J}}^F + L_{S_{X}}^{R^{lab+unl}} + L_{S_{X}}^F+ L_C^{R^{lab+unl}} + L_{DP}^{R^{lab}},
\end{displaymath}
where $lab$ denotes the \emph{labeled} soccer dataset and $unl$ denotes the \emph{unlabeled} CelebA dataset.

We treat $Y$ as a continuous variable to get around the issues of back-propagating through discrete variables. We also soften these values to range between $(-0.8, 0.8)$ and add stochasticity in the form of Gaussian noise with standard deviation $0.01$ for training stability. An alternative is to keep $Y$ discrete via the Gumbel-Softmax trick \cite{JangGP2017,MaddisonMT2017}, but given that it is just a single scalar variable alongside a high-dimensional $X$, it is simpler to treat $Y$ as continuous.

The classifiers for evaluation have the same architecture as the discriminator output $P(C\mid X)$. For stable GAN training, the  discriminator weights are regularized using spectral normalization \cite{MiyatoKKY2018}. Cross entropy loss is used for classification losses while hinge version of the adversarial loss is used for source (real or fake) losses \cite{2017arXiv170502894L}. We use Adam optimizer with hyperparameters set to $\beta_1=0.0$ and $\beta_2=0.9$. We also decay the learning linearly with the initial value set to $2e-4$.

We apply the Fairness GAN to the four datasets described in Section \ref{sec:intro}.  Details regarding the datasets are given in the supplement.  We perform a random 90/10 training/testing partition of the data (70/30 partition for soccer) and use the training set for independently learning two GANs: one with demographic parity loss and one with equality of opportunity loss.  Debiased datasets are generated from the learned generators.  

Then three separate classifiers are trained on the training partition of the original data without debiasing and the two generated datasets with debiasing.  Finally, these three separate classifiers are evaluated on the same testing partition of the original data.  The reported performance is based on an average across iterations of the classifier.  Because of the small number of real samples in the soccer dataset, it is impractical to train a classifier with a similar architecture to the GAN discriminator like we do with the other datasets. Therefore, we chose to only train the weights for the final fully-connected layer and use the weights from a classifier trained on CelebA faces for the previous layers.

\section{Empirical Results}
\label{sec:empirical}

Generally speaking, the demographic parity version of the Fairness GAN is able to improve the demographic parity and the equality of opportunity version of the Fairness GAN is able to improve the equality of opportunity.  We are unable to provide a comparative study with any other methods because this is the first pre-processing fairness method with treatment parity that can scale to high-dimensional multimedia features.

\begin{figure}
\centering
\begin{tabular}{cccc}
male = 0 & male = 0 & male = 1 & male = 1 \\
attractive = 0 & attractive = 1 & attractive = 0 & attractive = 1 \\
\includegraphics[width=0.2\textwidth]{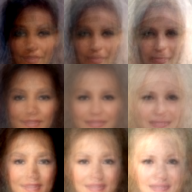} & \includegraphics[width=0.2\textwidth]{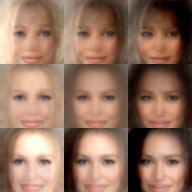} & \includegraphics[width=0.2\textwidth]{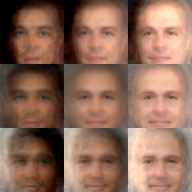} &
\includegraphics[width=0.2\textwidth]{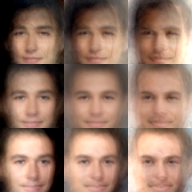} \\
\multicolumn{4}{c}{Without Debiasing} \\
\includegraphics[width=0.2\textwidth]{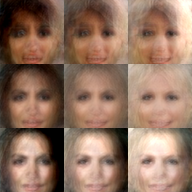} & \includegraphics[width=0.2\textwidth]{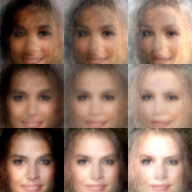} & \includegraphics[width=0.2\textwidth]{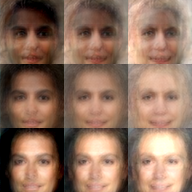} &
\includegraphics[width=0.2\textwidth]{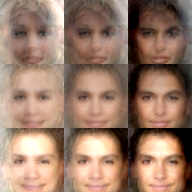} \\
\multicolumn{4}{c}{Fairness GAN DP} \\
\includegraphics[width=0.2\textwidth]{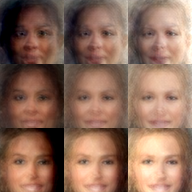} & \includegraphics[width=0.2\textwidth]{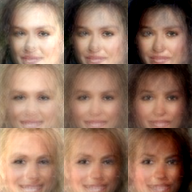} & \includegraphics[width=0.2\textwidth]{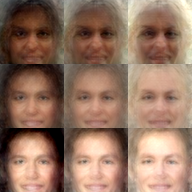} &
\includegraphics[width=0.2\textwidth]{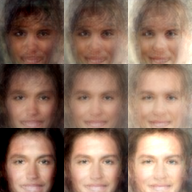} \\
\multicolumn{4}{c}{Fairness GAN Eq Opp}
\end{tabular}
\caption{Eigenfaces from the CelebA dataset (male, attractive).}
\label{fig:eigenfaces_celeba_male_attractive}
\end{figure}

\begin{figure}
\centering
\begin{tabular}{cccc}
dark & dark & light & light \\
attractive = 0 & attractive = 1 & attractive = 0 & attractive = 1 \\
\includegraphics[width=0.2\textwidth]{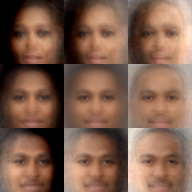} & \includegraphics[width=0.2\textwidth]{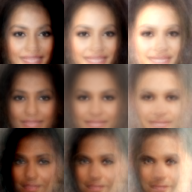} & \includegraphics[width=0.2\textwidth]{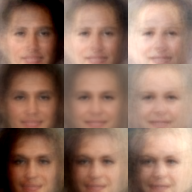} &
\includegraphics[width=0.2\textwidth]{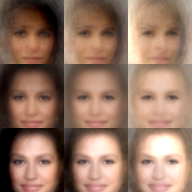} \\
\multicolumn{4}{c}{Without Debiasing} \\
\includegraphics[width=0.2\textwidth]{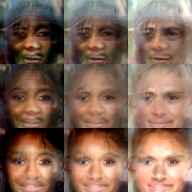} & \includegraphics[width=0.2\textwidth]{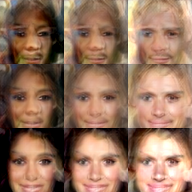} & \includegraphics[width=0.2\textwidth]{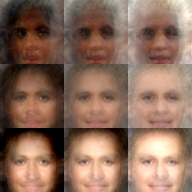} &
\includegraphics[width=0.2\textwidth]{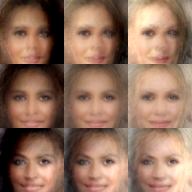} \\
\multicolumn{4}{c}{Fairness GAN DP} \\
\includegraphics[width=0.2\textwidth]{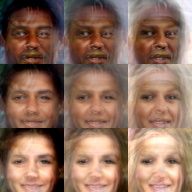} & \includegraphics[width=0.2\textwidth]{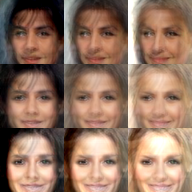} & \includegraphics[width=0.2\textwidth]{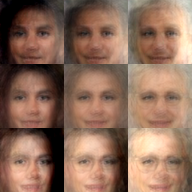} &
\includegraphics[width=0.2\textwidth]{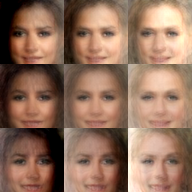} \\
\multicolumn{4}{c}{Fairness GAN Eq Opp}
\end{tabular}
\caption{Eigenfaces from the CelebA dataset (skin tone, attractive).}
\label{fig:eigenfaces_celeba_skintone_attractive}
\end{figure}

\begin{figure}
\centering
\begin{tabular}{cccc}
dark & dark & light & light \\
high card rate & low card rate & high card rate & low card rate \\
\includegraphics[width=0.2\textwidth]{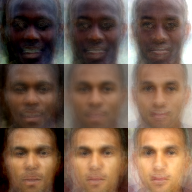} & \includegraphics[width=0.2\textwidth]{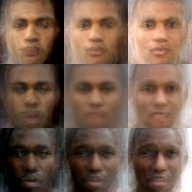} & \includegraphics[width=0.2\textwidth]{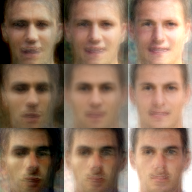} &
\includegraphics[width=0.2\textwidth]{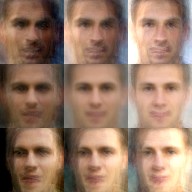} \\
\multicolumn{4}{c}{Without Debiasing} \\
\includegraphics[width=0.2\textwidth]{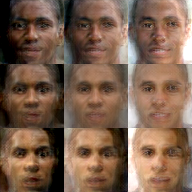} & \includegraphics[width=0.2\textwidth]{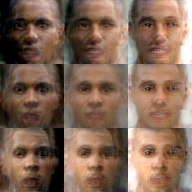} & \includegraphics[width=0.2\textwidth]{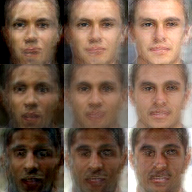} &
\includegraphics[width=0.2\textwidth]{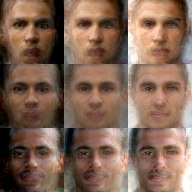} \\
\multicolumn{4}{c}{Fairness GAN DP} \\
\includegraphics[width=0.2\textwidth]{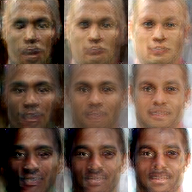} & \includegraphics[width=0.2\textwidth]{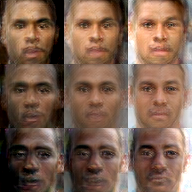} & \includegraphics[width=0.2\textwidth]{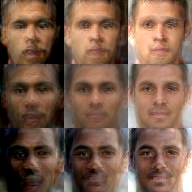} &
\includegraphics[width=0.2\textwidth]{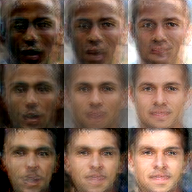} \\
\multicolumn{4}{c}{Fairness GAN Eq Opp}
\end{tabular}
\caption{Eigenfaces from the soccer dataset.}
\label{fig:eigenfaces_soccer}
\end{figure}

\begin{figure}
\centering
\begin{tabular}{cccc}
Great Britain & Great Britain & Canada & Canada \\
unrecognized & recognized & unrecognized & recognized \\
\includegraphics[width=0.2\textwidth]{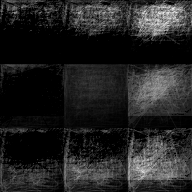} & \includegraphics[width=0.2\textwidth]{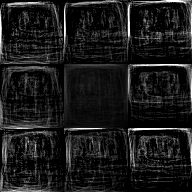} & \includegraphics[width=0.2\textwidth]{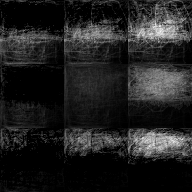} &
\includegraphics[width=0.2\textwidth]{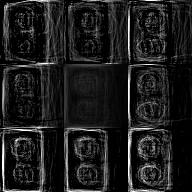} \\
\multicolumn{4}{c}{Without Debiasing} \\
\includegraphics[width=0.2\textwidth]{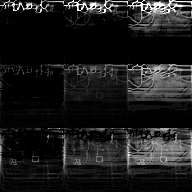} & \includegraphics[width=0.2\textwidth]{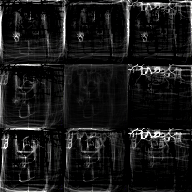} & \includegraphics[width=0.2\textwidth]{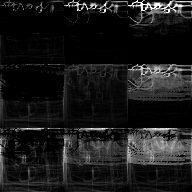} &
\includegraphics[width=0.2\textwidth]{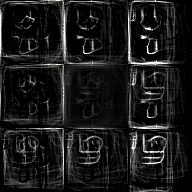} \\
\multicolumn{4}{c}{Fairness GAN DP} \\
\includegraphics[width=0.2\textwidth]{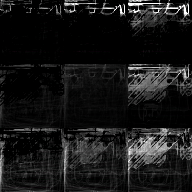} & \includegraphics[width=0.2\textwidth]{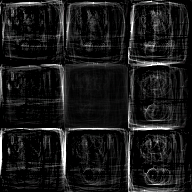} & \includegraphics[width=0.2\textwidth]{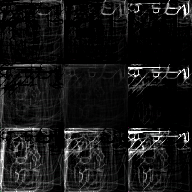} &
\includegraphics[width=0.2\textwidth]{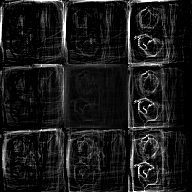} \\
\multicolumn{4}{c}{Fairness GAN Eq Opp}
\end{tabular}
\caption{Eigensketches from the Quick, Draw! dataset's power outlet category.}
\label{fig:eigensketches}
\end{figure}

Table \ref{table:errorrates} presents the different error rates for the three classifiers conditioned on the protected attribute, the overall unconditional error rate, and the values of demographic parity (difference of conditional error rates) and equality of opportunity (difference of conditional false negative rates) for all four datasets.
\begin{table}
\caption{Error rates, demographic parity, and equality of opportunity.}
\label{table:errorrates}
\centering
\begin{tabular}{c|c|c|c|c|c|c}
\toprule
\textbf{CelebA}  & \multicolumn{2}{c|}{Without Debiasing} & \multicolumn{2}{c|}{Fairness GAN DP} & \multicolumn{2}{c}{Fairness GAN Eq Opp} \\
\textbf{(male, attractive)} & male = 0 & male = 1 & male = 0 & male = 1 & male = 0 & male = 1 \\
\midrule
False Positive Rate & 0.4043 & 0.0927 & 0.5185 & 0.2356 & 0.4232 & 0.1672 \\ \hline 
False Negative Rate & 0.1213 & 0.4222 & 0.1821 & 0.4074 & 0.2119 & 0.4373 \\ \hline
Error Rate  & 0.2196 & 0.1749 & 0.2989 & 0.2785 & 0.2853 & 0.2346 \\ \hline
Unconditional Error Rate & \multicolumn{2}{c|}{0.2023} & \multicolumn{2}{c|}{0.2910} & \multicolumn{2}{c}{0.2657} \\ \hline
Demographic Parity & \multicolumn{2}{c|}{0.0447} & \multicolumn{2}{c|}{\textbf{0.0263}} & \multicolumn{2}{c}{0.0507} \\ \hline
Equality of Opportunity & \multicolumn{2}{c|}{0.2908} & \multicolumn{2}{c|}{0.2253} & \multicolumn{2}{c}{\textbf{0.2253}} \\
\bottomrule
\toprule
\textbf{CelebA} & \multicolumn{2}{c|}{Without Debiasing} & \multicolumn{2}{c|}{Fairness GAN DP} & \multicolumn{2}{c}{Fairness GAN Eq Opp} \\
\textbf{(skin tone, attractive)} & dark & light & dark & light & dark & light \\
\midrule
False Positive Rate & 0.1186 & 0.2035 & 0.3296 & 0.4761 & 0.2896 & 0.3755 \\ \hline 
False Negative Rate & 0.3099 & 0.1917 & 0.3279 & 0.2194 & 0.3652 & 0.2799 \\ \hline
Error Rate & 0.1846 & 0.1973 & 0.3290 & 0.3413 & 0.3157 & 0.3253 \\ \hline
Unconditional Error Rate & \multicolumn{2}{c|}{0.1945} & \multicolumn{2}{c|}{0.3394} & \multicolumn{2}{c}{0.3238} \\ \hline
Demographic Parity & \multicolumn{2}{c|}{0.0145} & \multicolumn{2}{c|}{\textbf{0.0136}} & \multicolumn{2}{c}{0.0197} \\ \hline
Equality of Opportunity & \multicolumn{2}{c|}{0.1162} & \multicolumn{2}{c|}{0.1085} & \multicolumn{2}{c}{\textbf{0.0853}} \\
\bottomrule
\toprule
\textbf{Soccer} & \multicolumn{2}{c|}{Without Debiasing} & \multicolumn{2}{c|}{Fairness GAN DP} & \multicolumn{2}{c}{Fairness GAN Eq Opp} \\
 & dark & light & dark & light & dark & light \\
\midrule
False Positive Rate & 0.2029 & 0.1428 & 0.1591 & 0.3466 & 0.3445 & 0.4814 \\ \hline 
False Negative Rate & 0.8202 & 0.8089 & 0.8850 & 0.6899 & 0.5492 & 0.5651 \\ \hline
Error Rate & 0.5459 & 0.4387 & 0.5624 & 0.4991 & 0.4582 & 0.5186 \\ \hline
Unconditional Error Rate & \multicolumn{2}{c|}{0.4602} & \multicolumn{2}{c|}{0.5118} & \multicolumn{2}{c}{0.5064} \\ \hline
Demographic Parity & \multicolumn{2}{c|}{0.1072} & \multicolumn{2}{c|}{0.0633} & \multicolumn{2}{c}{\textbf{0.0604}} \\ \hline
Equality of Opportunity & \multicolumn{2}{c|}{\textbf{0.0154}} & \multicolumn{2}{c|}{0.1950} & \multicolumn{2}{c}{0.0180} \\
\bottomrule
\toprule
\textbf{Quick, Draw!} & \multicolumn{2}{c|}{Without Debiasing} & \multicolumn{2}{c|}{Fairness GAN DP} & \multicolumn{2}{c}{Fairness GAN Eq Opp} \\
 & GB & CA & GB & CA & GB & CA \\
\midrule
False Positive Rate & 0.2638 & 0.3697 & 0.2951 & 0.3482 & 0.4213 & 0.4921 \\ \hline 
False Negative Rate & 0.0716 & 0.0189 & 0.1957 & 0.1343 & 0.0348 & 0.0111 \\ \hline
Error Rate & 0.1096 & 0.0509 & 0.2203 & 0.1565 & 0.1113 & 0.0549 \\ \hline
Unconditional Error Rate & \multicolumn{2}{c|}{0.0864} & \multicolumn{2}{c|}{0.1938} & \multicolumn{2}{c}{0.0890} \\ \hline
Demographic Parity & \multicolumn{2}{c|}{0.0587} & \multicolumn{2}{c|}{0.0639} & \multicolumn{2}{c}{\textbf{0.0563}} \\ \hline
Equality of Opportunity & \multicolumn{2}{c|}{0.0527} & \multicolumn{2}{c|}{0.0614} & \multicolumn{2}{c}{\textbf{0.0237}} \\
\bottomrule
\end{tabular}
\end{table}
\begin{figure}
  \centering
\begin{tabular}{cc}
\includegraphics[width=0.5\textwidth]{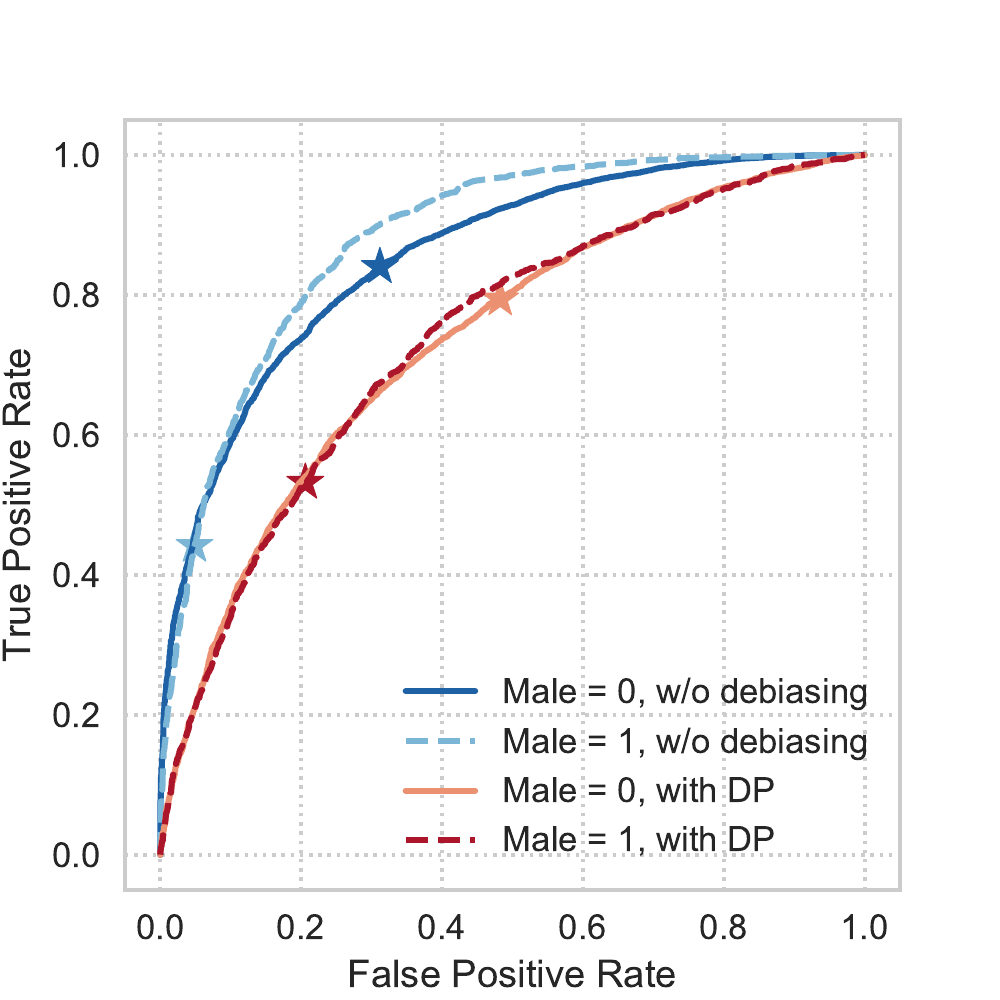} & \includegraphics[width=0.5\textwidth]{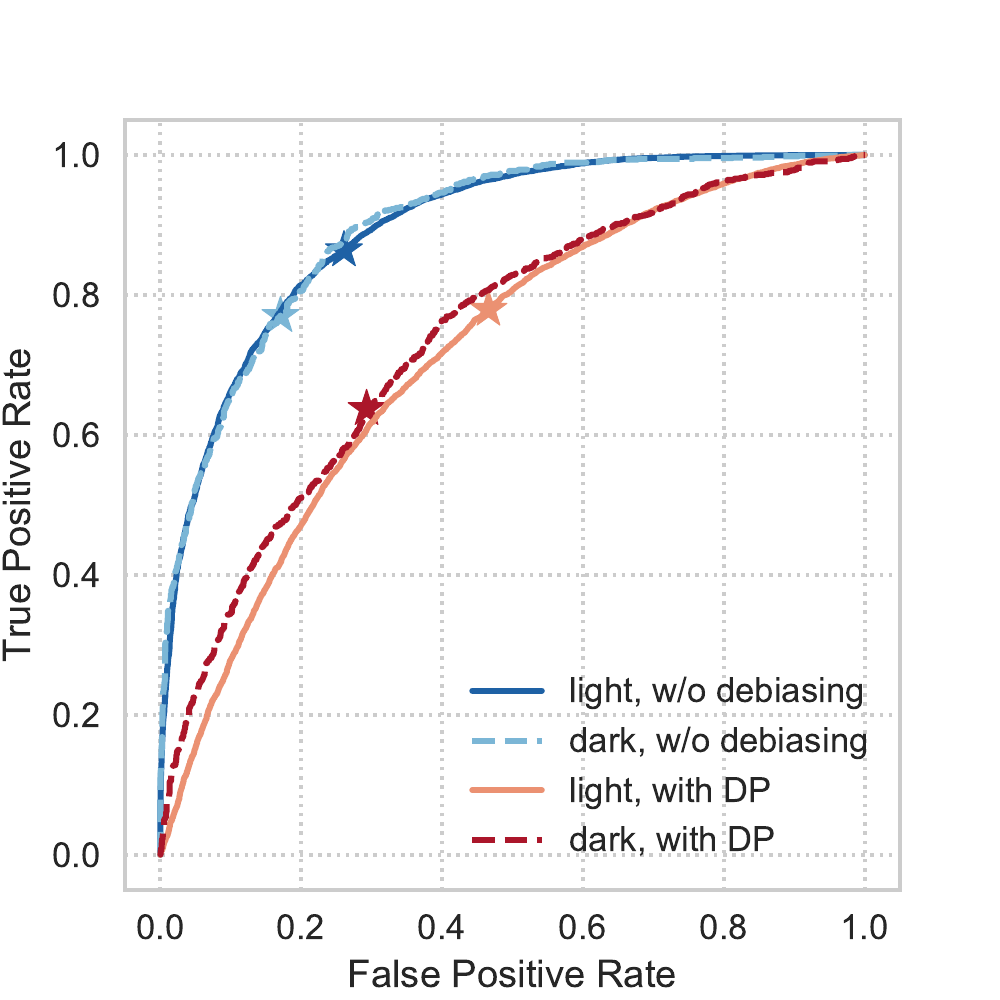} \\  
(a) & (b) \\
\includegraphics[width=0.5\textwidth]{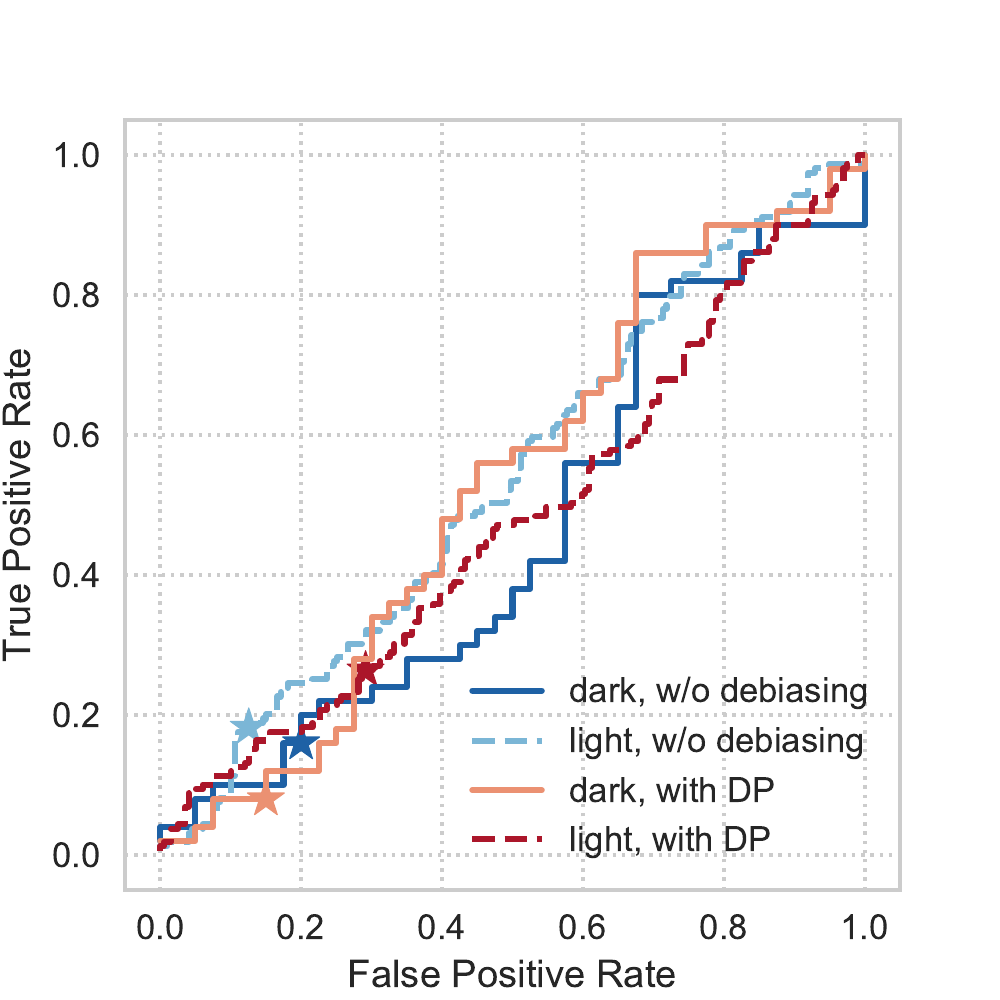} & \includegraphics[width=0.5\textwidth]{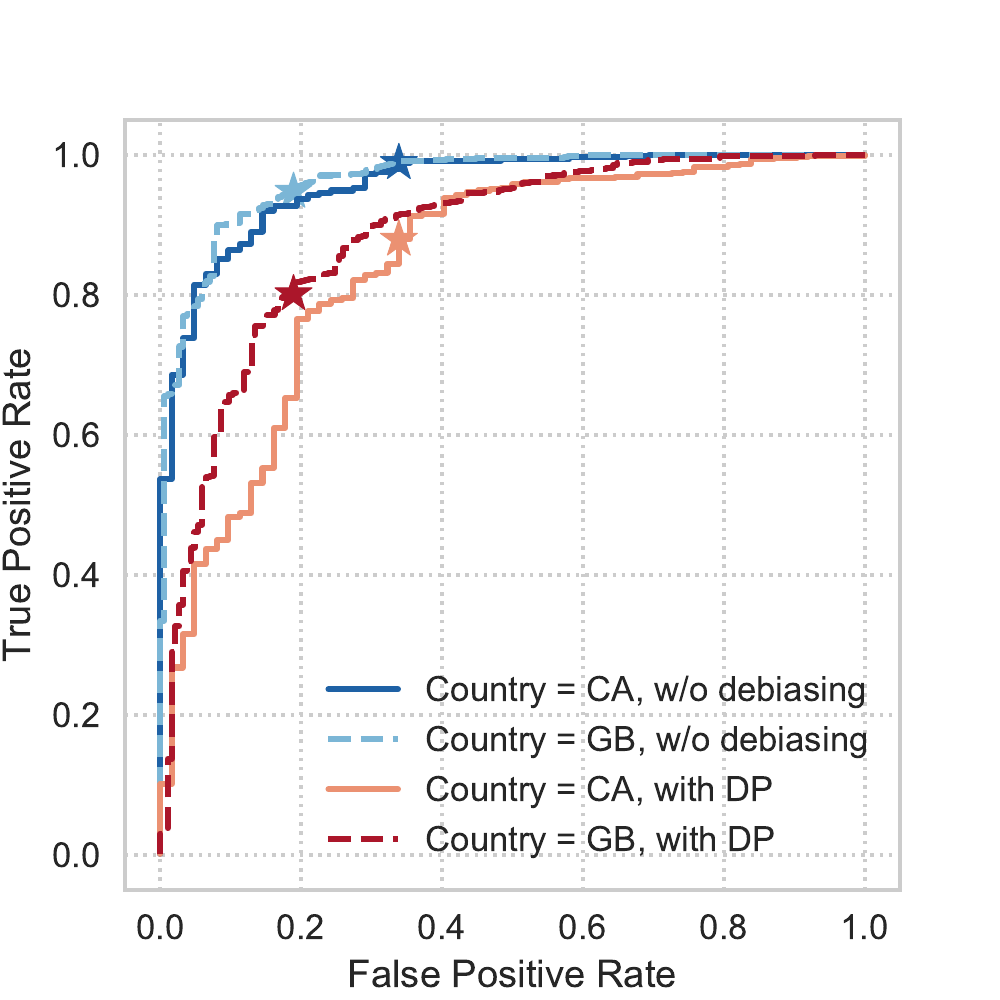} \\
(c) & (d) 
  \end{tabular}
  \caption{Receiver operating characteristics (left to right): CelebA (male, attractive), CelebA (skin tone, attractive), soccer, and Quick, Draw!. Marks show the threshold of 0.5 used by the classifier.}
  \label{fig:roc}
\end{figure}
Figure \ref{fig:roc} presents the receiver operating characteristics (ROCs) for the case without debiasing and the demographic parity Fairness GAN, broken down by the two values of the protected attribute.  The ROC of the equality of opportunity Fairness GAN is omitted to avoid cluttering the plots.  We visualize the image samples in Figure \ref{fig:eigenfaces_celeba_male_attractive} through Figure \ref{fig:eigensketches} as follows.  We compute the eigenfaces or eigensketches and show the mean image in the center of a 3 by 3 grid \cite{TurkP1991}.  The images to the left and right show variation along the first principal component.  The images to the top and bottom show variation along the second principal component.  The diagonal images show variation along both the first and second principal components together.

On CelebA (male, attractive), we see that debiasing takes the demographic parity value from 0.0447 to 0.0263 and the equality of opportunity from 0.2908 to 0.2253.  The overall accuracy does suffer in achieving these goals. A closer examination of the different types of errors shows that debiasing moves both the male and non-male operating points in the top and right direction of the ROC without changing their relative spacing. (A large change in relative spacing would require treatment disparity.) With debiasing, the false positive rate for non-males is quite high, which actually results in more favorable outcomes for this group.  This examination illustrates that different definitions of fairness act in different, perhaps unexpected, ways: equality of odds would yield even a further different result \cite{ZhangLM2018,KleinbergMR2017}.  Looking at the eigenfaces, we see that the demographic parity GAN makes both unattractive and attractive males presumably less attractive by femininizing their features: fuller lips, less defined jawline, and bigger eyes. The change caused by the equality of odds GAN is much less pronounced on the unattractive males, which makes sense because equality of odds is only concerned with $Y=1$. One of the desired properties of the Fairness GAN approach is to produce a realistic dataset that can be examined in a transparent way; our examination of lips and jawlines is exactly such a transparent examination not possible with latent representation based approaches.

CelebA (skin tone, attractive) already has excellent demographic parity, and the GAN improves it a bit.  The GAN improves the equality of opportunity from 0.1162 to 0.0853.  The eigenfaces show that the GANs equalize the skin tones across the groups. In fact, after debiasing for equality of opportunity, it appears that the dark attractive mean face has slightly lighter tone than the light attractive mean face.

The soccer dataset presents an inherently  difficult task because a face image is not a particularly useful feature to predict cautionable and sending off offenses.  Nonetheless, even for such a challenging dataset, the Fairness GAN does improve demographic parity quite a lot.  The equality of opportunity is already extremely small, and the equality of opportunity GAN maintains a small value; the demographic parity GAN, which does not have equality of opportunity in its objective causes it to degrade significantly\hspace{1pt}---\hspace{1pt}once again illustrating that the different fairness definitions behave in weird ways sometimes.  It is observed that the GANs serve to equalize skin tone in these eigenfaces as well. 

Finally, with the Quick, Draw!\ dataset, we notice that the GAN is not able to improve the demographic parity, but is able to improve the equality of opportunity from 0.0527 to 0.0237.  We suspect that too much distortion of individual images would be required to improve the demographic parity (a violation of individual fairness imposed implicitly in the GAN formalism).  The eigensketches are very interesting to examine.  The unrecognized sketches from both Great Britain and Canada have no structure to them but the recognized sketches do: more single socket square three-pronged patterns from Great Britain and double socket flat blade patterns from Canada (corresponding to those countries' respective power outlets).  After debiasing, the Great Britain eigensketches have more of the double socket character and the Canada ones have more of the single socket square character.

\section{Conclusion}
\label{sec:conclusion}

In this paper, we have examined allocative fairness in the scenario of binary classification with multimedia features and have developed a GAN-based pre-processing approach to improve demographic parity or equality of opportunity in the system by learning to generate a fairer dataset in the original input feature space. We use the proposed algorithm, the first application of GANs to algorithmic fairness, to process several attributed image datasets with varied properties, outcome variables, and protected attributes by adapting a unique combination of several recent techniques from the GAN literature.  The empirical results are generally positive, but do leave some room for improvement.

The work so far illuminates several directions for future research, e.g., considering other modalities of multimedia data in addition to images, adding the equality of odds fairness definition \cite{ZhangLM2018}, investigating the possible advantages to allowing treatment disparity, pursuing the Gumbel-Softmax trick for discrete outcome variables \cite{JangGP2017,MaddisonMT2017}, and formulating a Fairness GAN for continuous protected attributes \cite{LouppeKC2017}.


\bibliography{fairnessgan-nips}
\bibliographystyle{IEEEtran}

\end{document}